\documentclass{article}

\usepackage{arxiv}
\usepackage[utf8]{inputenc} % allow utf-8 input
\usepackage[T1]{fontenc}    % use 8-bit T1 fonts
\usepackage[colorlinks = true,
            linkcolor = blue,
            urlcolor  = blue,
            citecolor = blue,
            anchorcolor = blue]{hyperref}       % hyperlinks
\usepackage{url}            % simple URL typesetting
\usepackage{booktabs}       % professional-quality tables
\usepackage{amsfonts}       % blackboard math symbols
\usepackage{nicefrac}       % compact symbols for 1/2, etc.
\usepackage{microtype}      % microtypography
\usepackage{lipsum}
\usepackage{graphicx}
\usepackage{multicol}
\usepackage{float}

\title{Multi-agent Reinforcement Learning in OpenSpiel}

\author{
  Michael Walton \\
  Department of Computer Science\\
  Czech Technical University\\
  \texttt{waltomic@fel.cvut.cz} \\
   \And
 Viliam Lisy \\
  Department of Computer Science\\
  Czech Technical University\\
  \texttt{viliam.lisy@agents.fel.cvut.cz} \\
}

\begin{document}
\maketitle

\begin{abstract}
In this report, we present results reproductions for several core algorithms implemented in the OpenSpiel \cite{LanctotEtAl2019OpenSpiel} framework for learning in games. The primary contribution of this work is a validation of OpenSpiel's re-implemented search and Reinforcement Learning algorithms against the results reported in their respective originating works. Additionally, we provide complete documentation of hyperparameters and source code required to reproduce these experiments easily and exactly.
\end{abstract}

% keywords can be removed

\section{Introduction}
The OpenSpiel framework provides a collection of environments and algorithm implementations for studying Reinforcement Learning (RL) in games. OpenSpiel includes many popular general-sum, zero-sum, perfect and imperfect information games with episodic interfaces suitable for training RL agents. The algorithms implemented in OpenSpiel are contemporary or state-of-the-art (SOTA) and are designed to be highly configurable and extensible. As stated in the documentation and provided example code, the given default parameters are (in the majority of cases) intended to solve the imperfect information poker variant Kuhn \cite{kuhn2016contributions}. However, the papers originally proposing many of the OpenSpiel algorithms may not necissarily provide results for this environment and instead report results for more challenging games such as Leduc or Heads up No-Limit Texas Holdem. This limits OpenSpiel users' ability to convinently verify the correctness and performance of algorithim implementations using this tool. Therefore, it is interesting to validate that the partial results provided in the OpenSpiel paper \cite{LanctotEtAl2019OpenSpiel} can be reliably reproduced with comparable performance \footnote{For the purposes of this study, we aim to produce performances matching or exceeding those reported for their original implementations} to SOTA results present elsewhere in the literature. 

We have conducted a comprehensive reproduction study in which we extract relevant metrics and hyperparameters from SOTA publications and attempt to replicate these studies as faithfully and exactly as possible. The motivation of this effort is to rigorously validate the algorithms implemented in OpenSpiel as well as provide complete hyperparameter sets so that SOTA baselines may be quickly reproduced by other researchers. In the course of this study, we identified several experimental nuances and parameter sensitivities which we will discuss later in the paper. We hope that addressing these small, often undocumented concerns will prove helpful to the Multi-Agent RL community by providing insight and (perhaps more importantly) saving valuable time and effort.

\section{Quickstart \& Usage}
The source repository for reproducing these experiments can be found on github at \href{https://github.com/aicenter/openspiel_reproductions}{aicenter/openspiel\_reproductions}. An example OpenSpiel docker image can be pulled from Docker Hub at \href{https://hub.docker.com/repository/docker/waltonmyke/openspiel}{waltonmyke/openspiel} \footnote{We base our experiments on OpenSpiel release 0.2.0}. Individual experiments may be run using scripts in the \texttt{algorithms} foler. SLURM jobfiles can also be constructed and batch run with \texttt{run.py}.
Hyperparameter configurations to replicate the results in this paper can be found in the arguments defined in the \texttt{*.cfg} files in \texttt{config} and in Appendix \ref{appendix:hyperparams}. Finally, for methods that necissitated additional hyperparameter search, configurations of these procedures may be found under \texttt{sweep}.

We use \href{https://wandb.ai/site}{Weights \& Biases} \cite{wandb} for experiment configuration and results logging. If you prefer not to use this platform or do not have it configured for your environment, W\&B logging may be deactivated and redirected to standard out with the \texttt{-{}-no\_wandb} flag.

\section{Algorithms \& Methods}
While a rigorous technical survey of the methods utilized herein is outside the scope of this paper, we include high-level descriptions of each algorithm for proper attribution and to provide the reader with some context for relevant prior work. The reader may also refer to \cite{yang2020overview, Hernandez_Leal_2019} for more comprehensive surveys of recent techniques in Reinforcement Learning approaches to game solving.

\paragraph{Extensive Form Fictitious Play (XFP) \& Fictitious Self Play (FSP)} \cite{pmlr-v37-heinrich15} In classic fictitious play, players best respond to an average of their opponents' strategies on each iteration (or a slightly perturbed average of opponent strategies as in \cite{LESLIE2006285}). However, vanilla fictitious play is only defined in normal form (an exponentially less efficient representation for extensive form games). To address this \cite{pmlr-v37-heinrich15} proposed XFP which allows player strategy updates in the extensive form with linear time and space complexity. FSP is an approximation to XFP which replaces the best response computation with tabular reinforcement learning and estimates the average of opponent strategies using supervised learning over action probabilities.

\paragraph{Regression CFR (RCFR)} \cite{waugh2014solving} estimates counterfactual regret using online updates to a function approximator and improves it's strategy by minimizing the predicted regret of the estimator on each iteration. In the original work the authors proposed regression trees as a function approximator and empirically evaluate the impact of error tolerance at the leaf nodes on the quality of the strategies obtained. In OpenSpiel, RCFR regrets are estimated using a neural network.

\paragraph{DeepCFR} \cite{brown2019deep} Similar to RCFR, DeepCFR uses function approximation to generalize across similar infosets obviating the need to calculate and accumulate regrets for each infoset. DeepCFR uses external sampling Monte Carlo CFR (MCCFR) to conduct a number of partial traversals of the game tree and update the values of states along the trajectory on each iteration. The observed values are stored in a memory buffer and used to update the parameters of a \emph{value prediction} network which estimates the value of an action in an infostate. Because the \emph{average} strategy of all CFR iterations converges to a Nash Equilibrium, it is insufficient to maintain a single parametric policy. To address this, DeepCFR maintains a separate \emph{policy network} which approximates the average strategy by learning a distribution over actions selected for each infostate on each iteration.

\paragraph{Exploitability Descent (ED)} \cite{DBLP:journals/corr/abs-1903-05614} directly updates the player's policy against a worst case opponent in a two-player zero sum game. The exploitability of each player employing ED's strategy converges asymptotically to zero; hence in self-play, the joint strategy converges to an approximate Nash Equilibrium. On each iteration, ED first computes a best response to each players' policy. Next, update the policy using gradient ascent to maximize player expected utility wrt. to the best response of the other player.

\paragraph{Neural Fictitious Self Play (NFSP)} \cite{DBLP:journals/corr/HeinrichS16} augments FSP with deep learning by replacing the tabular Reinforcement Learning agent with a neural network (trained off-policy with $\epsilon$-greedy Q-learning) and the (infostate, action)-pair visitation counts with an additional policy network which approximates average strategies by maximizing the log-probability of observed actions in each infostate. NFSP also incorporates a short-horizon prediction of opponent strategies using \emph{anticipatory dynamics} for which the authors demonstrate improved empirical performance.

\paragraph{Policy Space Response Oracles (PSRO)} \cite{lanctot2017unified} generalizes fictitious play and the well-studied Double Oracle algorithm \cite{10.5555/3041838.3041906}. In general PSRO, an empirical game (much smaller in size than the original game) is incrementally constructed by playing the full game with a set of policies sampled from a portfolio. A meta-strategy is induced over a resultant empirical payoff table which is estimated by rolling out strategies currently in the portfolio. New policies are discovered by iteravely setting one player as the fixed-strategy oracle and training a new strategy as a best response. In general, a specific PSRO-family algorithm can be (principally) characterized by a particular choice of meta-strategy solver and oracle.

\paragraph{Policy Gradient} \cite{DBLP:journals/corr/abs-1810-09026}, in the context of games, refers to a family of algorithms utilizing the policy gradient theorem from reinforcement learning \cite{pgsutton}. More specifically, variations on Advantage Actor-Critic (A2C) \cite{DBLP:journals/corr/MnihBMGLHSK16} in which an advantage function is used to reduce the variance of estimates of the policy gradient. Several variations on this algorithm are evaluated in the original work and implemented in OpenSpiel: \emph{Q-based Policy Gradient} uses the action-values of all actions rather than only those executed during a rollout to estimate the policy gradient. Hence it is also referred to as Mean Actor-Critic (MAC) in the literature \cite{allen2018mean}, \emph{Regret Policy Gradient} takes inspiration from CFR and represents the loss in terms of regret represented as a relu-thresholded difference between action-values and the baseline. \emph{Regret Matching Policy Gradient} extends this further by weighting the A2C policy gradient by the thresholded regrets.

\paragraph{Neural Replicator Dynamics (NeuRD)} \cite{hennes2020neural} is another policy gradient algorithm which yields desirable theoretical properties and greatly improved empirical performance through a slight modification of vanilla softmax PG. NeuRD also has formal relationships to the Hedge algorithm and sofmax-CFR; the latter of which guarantees convergence in the tabular case.

\paragraph{Metrics}
In cases where exact quantitative performance tables were not readily available or we would like to compare against convergence plots we extract estimated values from figures using \href{https://automeris.io/WebPlotDigitizer/}{webplotdigitizer} \footnote{Most original works present convergence plots on log-log scale. In the case where they are instead given on a linear scale, we present our replications in both linear and log. The aim is to clarify small differences between various parameters of each algorithm. However, note that small errors in the target curve traces extracted on linear scale may be exaggerated}. Original works may report results and figures using Exploitability or NashConv, possibly expressed in terms of milliblinds per hand (mb/h) in the case of poker. For consistency, and as we only report results for two-player zero-sum games (more specifically poker variants) we measure the performance of all strategy profiles in terms of Exploitability following the definitions in equations \ref{eqn:br} - \ref{eqn:exp}.

A Best Response $BR$ to a strategy profile including all players other than player $i$, denoted $\pi_{-i}$ is the set of strategies for a player $i$ which maximise their payoff function $u_i$ given the other players play $\pi_{-i}$

\begin{equation}
\label{eqn:br}
    B R\left(\pi_{-i}\right)=\left\{\pi_{i}^{\prime} \mid \pi_{i}^{\prime} = {argmax}_{u_{i}}\left(\pi_{i}, \pi_{-i}\right)\right\}
\end{equation}

The $NashConv$ of a strategy profile $\pi$ is defined as the sum over players $\mathcal{N}$ of each of their respective incentives to deviate from their current strategy to a best response, denoted $\delta_i(\pi)=u_i(\pi_i^b, \pi_{-i})-u_{i}(\pi)$ where $\pi^b_i \in BR(\pi_{-i})$

\begin{equation}
    \mathrm{NashConv}(\pi)=\sum_{i \in \mathcal{N}} \delta_{i}(\pi)
\end{equation}

When $|\mathcal{N}| = 2$, the exploitability of a strategy profile $\pi$ is

\begin{equation}
\label{eqn:exp}
    \mathrm{ Exploitability }(\pi)=\frac{\mathrm{NashConv}(\pi)}{|\mathcal{N}|}=\frac{\sum_{i \in \mathcal{N}} \delta_{i}(\pi)}{n}=\frac{u_{1}\left(\pi_{1}^{b}, \pi_{2}\right)+u_{2}\left(\pi_{1}, \pi_{2}^{b}\right)}{2}
\end{equation}

\section{Replications}
In this section we will summarize at a high-level the specific methods used to obtain each result. Unless otherwise specified, all hyperparameter search was conducted using the Hyperband \cite{DBLP:journals/corr/LiJDRT16} implementation provided in \cite{wandb}. Hyperband is a bandit-based hyperparameter search algorithm which utilizes adaptive resource allocation and configurable early stopping criteria to speed up traditional hyperparameter search based on adaptive configuration selection with Bayesian optimization. In our experiments, we set the total number of brackets $s=2$.

\begin{figure}[h]
    \centering
    \includegraphics[width=\textwidth]{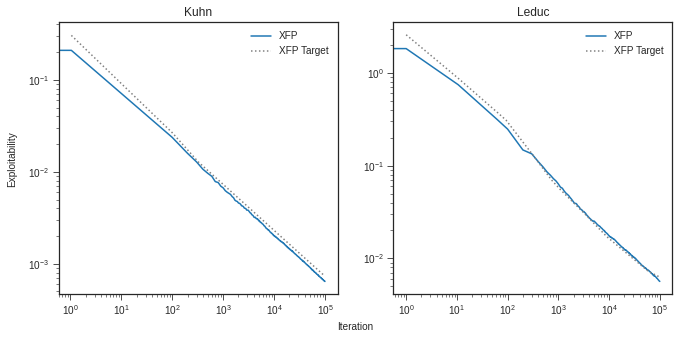}
\caption{Extensive Form Fictitious Play. \textit{XFP Target} from \cite{pmlr-v37-heinrich15} \label{figure:xfp}}
\end{figure}

Reproducing the \textbf{CFR \& XFP} results using their respective OpenSpiel implementations is relatively trivial using the default configuration implemented in the provided examples (Figure \ref{figure:xfp}). It is suggested, however, to use these reproductions as smoke-tests and sanity-checks for novel extensions to OpenSpiel (for instance the addition of new games) as their behavior is more predictable than more complex or harder to tune models.

\begin{figure}[h]
    \centering
    \includegraphics[width=\textwidth]{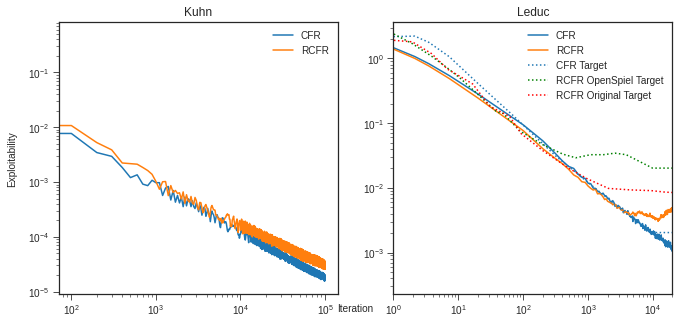}
\caption{Comparison of RCFR \& tabular CFR on Kuhn (left). Replication of results in Leduc (right). \textit{CFR Target} and \textit{RCFR OpenSpiel Target} extracted from \cite{LanctotEtAl2019OpenSpiel} \textit{RCFR Original Target} from \cite{waugh2014solving} \label{figure:rcfr}}
\end{figure}

\paragraph{RCFR} Although the high-level RCFR algorithm implemented in OpenSpiel aligns with the original work proposing RCFR, some implementation details differ. Most importantly, the results in the original work used regression trees to estimate regrets whereas OpenSpiel uses neural network estimator. We further observed that the best result reported for the original RCFR implementation exceeded the performance reported in OpenSpiel. Using the network size reported in the OpenSpiel paper (400 hidden units, two layer fully-connected), default batch size and number of epochs specified in the example code (200 and 100 respectively). These parameters yielded roughly approximate results to OpenSpiel. To match the performance of the original RCFR, we swept the learning rate over ${.1, .01, .001}$ and found .001 allowed the model to exceed the performance of the original regression tree based RCFR implementation, as demonstrated in Figure \ref{figure:rcfr}.

\begin{figure}[h]
    \centering
    \includegraphics[width=\textwidth]{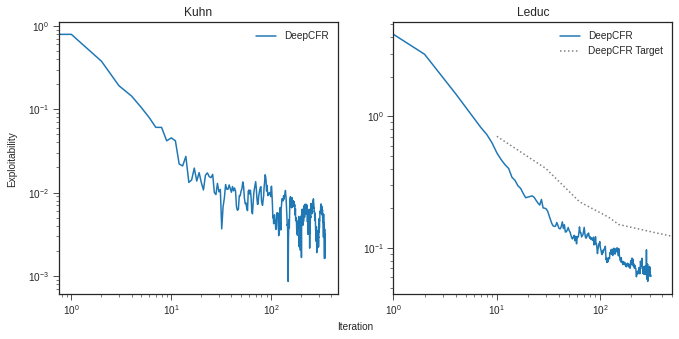}
\caption{DeepCFR. Target results from \cite{brown2019deep} \label{figure:deepcfr}}
\end{figure}

\paragraph{DeepCFR} We provide results comparisons against \cite{DBLP:journals/corr/abs-1901-07621} as this work demonstrated improved results for this algorithm over \cite{brown2019deep}. We used the suggested parameters from \cite{brown2019deep} however we also conducted a parameter search over the number of hidden units \{64, 128, 256\} the number of layers 3-5 and the number of external sampling iterations per iteration \{15k, 30k, 100k\}. As in \cite{brown2019deep} we re-initialized the value networks, trained the advantage network and policy network for 750 and 5000 gradient steps per iteration with a learning rate of $10^-3$ and minibatches of size 2048. As shown in Figure \ref{figure:deepcfr}, we find that the implementation in OpenSpiel, using these parameters, improves slightly over the target baseline.

\begin{figure}[h]
    \centering
    \includegraphics[width=\textwidth]{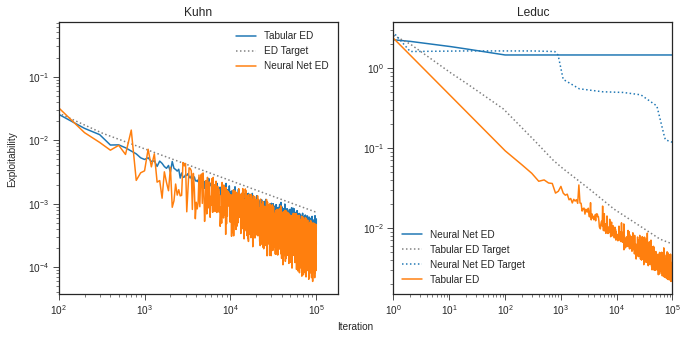}
\caption{Exploitability Descent. Target results \textit{Tabular ED} and \textit{Neural Net ED} in both domains from \cite{DBLP:journals/corr/abs-1903-05614} \label{figure:ed}}
\end{figure}

\paragraph{Exploitability Descent} Our ED replications in both the tabular and approximate implementations closely match the results presented in \cite{DBLP:journals/corr/abs-1903-05614} in the Kuhn domain. \footnote{The implementation in OpenSpiel was found to have a (very minor) bug in the provided example code in release 0.2.0; this is patched in our repository as well as OpenSpiel commit \href{https://github.com/deepmind/open_spiel/commit/291c5e241cc4744201c67607dcd4c20334189c37}{291c5e2}.} We conducted parameter sweeps over the initial learning rate for powers of 2 and (in the neural network case) number of hidden layers 1-5, number of hidden units $\{64,128,256\}$ and regularization weights from $[10^{-7}, 10^{-4}]$. We found that a three layer network with 256 hidden, and a regularization weight of .001 consistently outperformed other configurations, illustrated in Figure \ref{figure:ed}, and exhibited similar empirical convergence rates across multiple random seeds.

The originating study \cite{DBLP:journals/corr/abs-1903-05614} also presents results for more complex domains such as Leduc and Goofspiel. However, in our experiments we were unable to reproduce the results presented using neural networks. The same parameter searches were conducted for each domain as they were in Kuhn. We observed that although the tabular model converged as expected, neural network estimators demonstrated inconsistent behavior and occasionally failed to improve at all over the number of iterations tested. We expect that more reliable convergence could be accomplished by taking several gradient steps on the for each iteration (i.e. without updating the best response for each iteration of gradient descent). This could be accomplished by modifying the OpenSpiel implementation to step the minimizer on the advantage and policy losses without re-computing a new best-response to the modified strategy for some variable number of timesteps. While this change is relatively simple, we defer modifications to OpenSpiel's ED implementation for future work.

\begin{figure}[h]
    \centering
    \includegraphics[width=\textwidth]{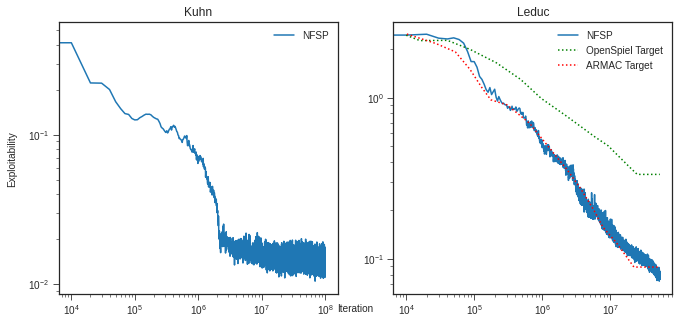}
\caption{Neural Fictitious Self Play. \textit{OpenSpiel Target} indicates results reported in \cite{LanctotEtAl2019OpenSpiel} \textit{ARMAC Target} trace extracted from results in \cite{armac} \label{figure:nfsp}}
\end{figure}

\paragraph{NFSP} Instead of the default parameters provided with OpenSpiel, we use a similar configuration for the NFSP baseline reported in the recently proposed \textit{Advantage Regret Matching Actor Critic} (ARMAC) \cite{armac} which demonstrates improved performance over the originaly reported OpenSpiel baseline. Specifically, we use the same network architecture with four fully-connected layers each with 128 hidden units. The anticipatory parameter $\eta=0.1$, the exploration parameter $\epsilon$ is decayed during the first 20m iterations, the reservoir and replay buffer capacities are limited to 2m entries and 200k respectively. We then conducted search over the following hyperparameters using the method previously described: $\epsilon$-greedy starting values \{0.06, 0.24\}, RL learning rates \{0.1, 0.01, 0.001\}, SL learning rates \{0.01, 0.001, 0.005), and DQN target network update period of \{1000, 19200\}. We identified and report results using the following configuration for leduc: $\epsilon$-greedy starting value 0.6, RL learning rate 0.01, and update period 19200. As illustrated in Figure \ref{figure:nfsp}, our results using these parameters match those reported in ARMAC very closely.

\begin{figure}[h]
    \centering
    \includegraphics[width=\textwidth]{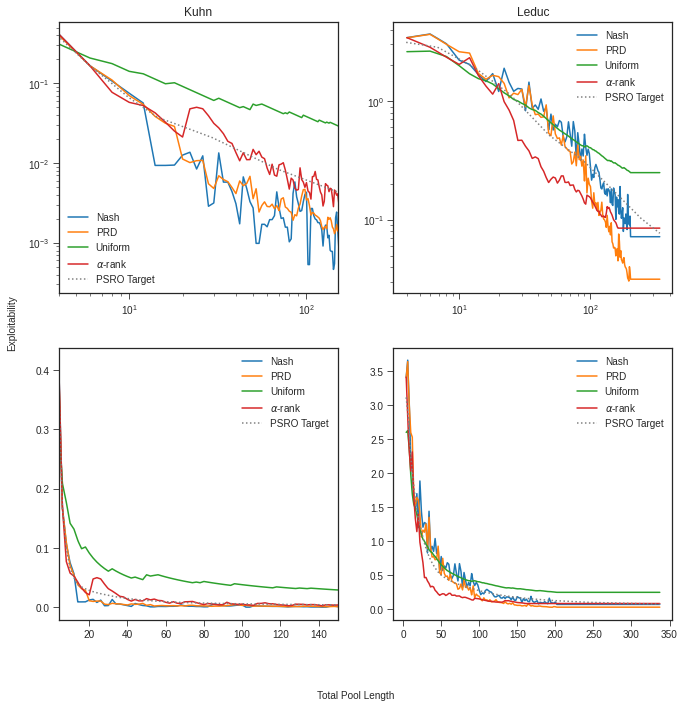}
\caption{Policy Space Response Oracles for different choices of meta-solver: \textit{Nash, PRD} (Projected Replicator Dynamics), \textit{Uniform} and $\alpha$- \textit{rank}. Target is extracted from \cite{DBLP:journals/corr/abs-1909-12823}. Note that the x-axis is the \textit{Total Pool Length} rather than iteration. \label{figure:psro}}
\end{figure}

\paragraph{PSRO} We defer description of the hyperparmeters for PSRO in Appendix \ref{appendix:hyperparams} as they are more numerous than other evaluated algorithms and thus, likely easier for the reader to parse in tabular form. Our experiments used the hyperpareter sweeps suggested in \cite{DBLP:journals/corr/abs-1909-12823} and followed a similar experimental procedure. As in \cite{DBLP:journals/corr/abs-1909-12823}, we elect to measure exploitability in terms of \textit{Total Pool Length} (the sum of the sizes of all players' policy sets) rather than iteration. This preference is motivated in the original work to encourage fair comparison across meta-solvers as some can add more than one strategy to a player's profile on a single iteration. As illustrated in Figure \ref{figure:psro}, our results for different metasolver choices track the baseline target from \cite{DBLP:journals/corr/abs-1909-12823}, however they do not match exactly for each \textit{choice} of solver. It is likely that additional parameter sweeps for each solver independently could improve this correspondence, however for simplicity we leave this to the reader who may have specific motivations to more rigorously compare the properties of different metagame solvers in PSRO.

\begin{figure}[h]
    \centering
    \includegraphics[width=\textwidth]{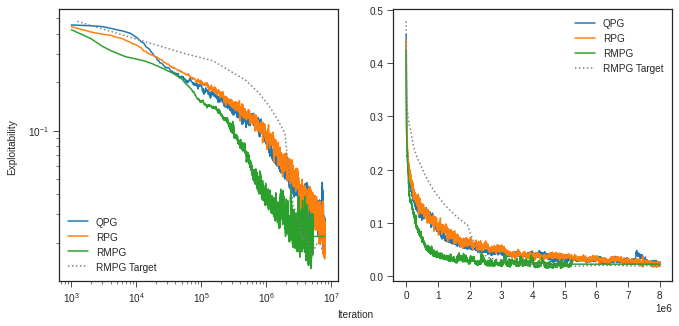}
\caption{Policy Gradients. Our replication of \textit{QPG, RPG \& RMPG} in the Kuhn domain vs. the RMPG results presented in \cite{DBLP:journals/corr/abs-1810-09026} \label{figure:pg}}
\end{figure}

\paragraph{Policy Gradients} In our policy gradient experiments (QPG, RPG, RMPG) we conducted similar parameter sweeps to those specified in \cite{DBLP:journals/corr/abs-1810-09026}: the learning rates for the policy and critic networks were swept over $\{.001, .01, .1\}$ independently for both networks. The number of hidden layers and hidden units ranged from 1-5 and $\{64,128,256\}$ respectively. The entropy cost was swept over $\{0, .01, .1, .15, .2\}$. The batch size and $N_q$ (the number of critic updates before a policy update) were selected from powers of 2 ranging 4-32 and 16-512 respectively. Although we were able to achieve comparable performance to the original study, the rates of convergence differed slightly in our replications. Interestingly, our results specifically improved the relative performance of RMPG (which in the prior work under-performed in Kuhn) over other policy gradient methods. In our experiments, we observed the convergence of all policy gradient methods evaluated strongly depend on the number of critic updates before updating the policy. This makes sense as improved estimates of the advantage function (regret in the case of RPG, RMPG) reduce the variance of the estimated policy gradient leading to smoother convergence. As illustrated in Figure \ref{figure:pg}, we were not able to quantitatively reproduce the target learning curves, however the OpenSpiel implementation demonstrates some modestly improved sample complexity early in training and converges to a comparable exploitability to \cite{DBLP:journals/corr/abs-1810-09026} after $10^7$ iterations.

\begin{figure}[h]
    \centering
    \includegraphics[width=\textwidth]{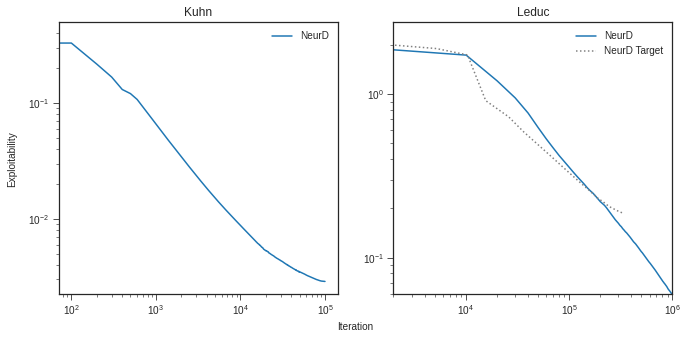}
\caption{Neural Replicator Dynamics. \textit{NeurD Target} from \cite{hennes2020neural} \label{figure:neurd}}
\end{figure}

\paragraph{Neural Replicator Dynamics} We observed that the default parameters in OpenSpiel examples and unit tests do not perform particularly well in the Leduc domain. Therefore, we conducted a hyperparameter search similar to the procedure described in \cite{hennes2020neural}. Our results substantially improve over those provided in OpenSpiel, however they do not quite match those reported in the original NeurD paper. We swept the number of hidden units and depth of the network over \{64, 128, 256\} and \{2,3,4\} layers respectively. The step size was sampled from \{.5, .9, 1., 1.5, 2., 2.5, 3., 3.5, 4.\} we also varied the logit threshold parameter from 1-3 (although the original work set the threshold to 2, we observed a slightly higher threshold to improve performance in our experiments). In a practical implementation of NeurD, the threshold parameter is used to improve the numerical stability of the algorithm during training. In our experiments, parameter searches consistently resulted in a preference for higher threshold values. As illustrated in Figure \ref{figure:neurd}, OpenSpiel's implementation \footnote{There is a small discrepancy between the threshold implementation described in original work and the OpenSpiel implementation. For the interested reader, \href{https://github.com/deepmind/open_spiel/issues/148}{issue 148} addresses this concern} matches the original work closely.

\section{Conclusion}
In this study, we have conducted a thorough battery of replications for the multi-agent reinforcement learning algorithms implemented in OpenSpiel. Where possible, we utilized network architectures and hyperparameter settings suggested elsewhere in the literature. We also conducted extensive parameter sweeps in settings where configurations suggested in prior work was insufficient to achieve expected behaviors of the evaluated algorithms. In the majority of cases, we were able to match or exceed the best-reported performance for each algorithm in OpenSpiel. As there are still several cases we were unable to address in this study, we intend to continue maintaining the provided source-code as well as this document if the expected behavior for current failure-cases can be accomplished. To this ends, we invite other OpenSpiel users to contribute if, in the course of their research, they address one of the limitations of this study. This may take the form of pull-requests on the associated git repository or other experimental documentation sufficient for exact reproduction. It is our hope that others working on reinforcement learning in games will benefit from this effort as a quick reference and a tool for improved reproducibility of their experiments.

\bibliographystyle{unsrt}  
\bibliography{references}  %%% Remove comment to use the external .bib

\appendix
\section{Hyperparameters}
\label{appendix:hyperparams}
Hyperparameter configurations for each algorithm for each game may be found in the \texttt{*.cfg} files in the \texttt{config/} folder. We provide them here as well for ease of reference.

\begin{multicols}{2}

\begin{table}[H][H]
\centering
\caption{qpg}
\begin{tabular}{ll}
\toprule
  Parameter & Value \\
\midrule
 num\_hidden &   128 \\
 num\_layers &     2 \\
\bottomrule
\end{tabular}
\end{table}
 \begin{table}[H]
\centering
\caption{exp\_descent}
\begin{tabular}{ll}
\toprule
Parameter & Value \\
\midrule
  init\_lr &   1.0 \\
 lr\_scale &   .01 \\
\bottomrule
\end{tabular}
\end{table}
 \begin{table}[H]
\centering
\caption{nn\_exp\_descent}
\begin{tabular}{ll}
\toprule
         Parameter &  Value \\
\midrule
           init\_lr &      1 \\
          lr\_scale &    .01 \\
 regularizer\_scale &  0.001 \\
        num\_hidden &    256 \\
        num\_layers &      3 \\
\bottomrule
\end{tabular}
\end{table}
 \begin{table}[H]
\centering
\caption{rpg}
\begin{tabular}{ll}
\toprule
            Parameter & Value \\
\midrule
           num\_hidden &   128 \\
           num\_layers &     2 \\
           batch\_size &     4 \\
         entropy\_cost &   0.1 \\
 critic\_learning\_rate &  0.01 \\
     pi\_learning\_rate &  0.01 \\
 num\_critic\_before\_pi &   128 \\
\bottomrule
\end{tabular}
\end{table}
 \begin{table}[H]
\centering
\caption{neurd}
\begin{tabular}{ll}
\toprule
            Parameter &  Value \\
\midrule
    num\_hidden\_layers &      2 \\
     num\_hidden\_units &    128 \\
   num\_hidden\_factors &      0 \\
 use\_skip\_connections &   True \\
           batch\_size &    100 \\
            threshold &     2. \\
            step\_size &    1.0 \\
           autoencode &  False \\
\bottomrule
\end{tabular}
\end{table}
 \begin{table}[H]
\centering
\caption{rcfr}
\begin{tabular}{ll}
\toprule
            Parameter &  Value \\
\midrule
            bootstrap &  False \\
    truncate\_negative &  False \\
          buffer\_size &     -1 \\
    num\_hidden\_layers &      2 \\
     num\_hidden\_units &    400 \\
   num\_hidden\_factors &      0 \\
 use\_skip\_connections &   True \\
           num\_epochs &    200 \\
           batch\_size &    100 \\
            step\_size &  0.001 \\
\bottomrule
\end{tabular}
\end{table}
 \begin{table}[H]
\centering
\caption{deep\_cfr}
\begin{tabular}{ll}
\toprule
                       Parameter &    Value \\
\midrule
                  num\_traversals &     1500 \\
            batch\_size\_advantage &     2048 \\
             batch\_size\_strategy &     2048 \\
                      num\_hidden &       64 \\
                      num\_layers &        3 \\
 reinitialize\_advantage\_networks &     True \\
                   learning\_rate &     1e-3 \\
                 memory\_capacity &  1000000 \\
      policy\_network\_train\_steps &     5000 \\
   advantage\_network\_train\_steps &      750 \\
\bottomrule
\end{tabular}
\end{table}
 \begin{table}[H]
\centering
\caption{nfsp}
\begin{tabular}{ll}
\toprule
                   Parameter &            Value \\
\midrule
                  eval\_every &            10000 \\
         hidden\_layers\_sizes &  128,128,128,128 \\
      replay\_buffer\_capacity &           200000 \\
   reservoir\_buffer\_capacity &          2000000 \\
    min\_buffer\_size\_to\_learn &             1000 \\
          anticipatory\_param &               .1 \\
                  batch\_size &              128 \\
                 learn\_every &              128 \\
            rl\_learning\_rate &              .01 \\
            sl\_learning\_rate &              .01 \\
               optimizer\_str &              sgd \\
 update\_target\_network\_every &            19200 \\
             discount\_factor &              1.0 \\
      epsilon\_decay\_duration &         20000000 \\
               epsilon\_start &              .06 \\
                 epsilon\_end &             .001 \\
           evaluation\_metric &   exploitability \\
\bottomrule
\end{tabular}
\end{table}
 \begin{table}[H]
\centering
\caption{psro}
\begin{tabular}{ll}
\toprule
                   Parameter &          Value \\
\midrule
                   n\_players &              2 \\
        meta\_strategy\_method &           nash \\
    number\_policies\_selected &              1 \\
              sims\_per\_entry &           1000 \\
            gpsro\_iterations &            100 \\
              symmetric\_game &          False \\
              prd\_iterations &          50000 \\
  training\_strategy\_selector &  probabilistic \\
                 oracle\_type &             BR \\
    number\_training\_episodes &          10000 \\
        self\_play\_proportion &            0.0 \\
           hidden\_layer\_size &            256 \\
                  batch\_size &             32 \\
                       sigma &            0.0 \\
               optimizer\_str &           adam \\
             num\_q\_before\_pi &              8 \\
             n\_hidden\_layers &              4 \\
                entropy\_cost &          0.001 \\
        critic\_learning\_rate &           0.01 \\
            pi\_learning\_rate &          0.001 \\
           dqn\_learning\_rate &           0.01 \\
 update\_target\_network\_every &           1000 \\
                 learn\_every &             10 \\
                        seed &              1 \\
                     verbose &           True \\
\bottomrule
\end{tabular}
\end{table}

\end{multicols}

\end{document}